# Medical Image Fusion: A survey of the state of the art

A. P. James[a], B. V. Dasarathy[b]

[a] Nazarbayev University, Email: apj@ieee.org

[b] Information Fusion Consultant

## Abstract

Medical image fusion is the process of registering and combining multiple images from single or multiple imaging modalities to improve the imaging quality and reduce randomness and redundancy in order to increase the clinical applicability of medical images for diagnosis and assessment of medical problems. Multi-modal medical image fusion algorithms and devices have shown notable achievements in improving clinical accuracy of decisions based on medical images. This review article provides a factual listing of methods and summarizes the broad scientific challenges faced in the field of medical image fusion. We characterize the medical image fusion research based on (1) the widely used image fusion methods, (2) imaging modalities, and (3) imaging of organs that are under study. This review concludes that even though there exists several open ended technological and scientific challenges, the fusion of medical images has proved to be useful for advancing the clinical reliability of using medical imaging for medical diagnostics and analysis, and is a scientific discipline that has the potential to significantly grow in the coming years.

## 1. Introduction

Medical image fusion encompasses a broad range of techniques from image fusion and general information fusion to address medical issues reflected through images of human body, organs, and cells. There is a growing interest and application of the imaging technologies in the areas of medical diagnostics, analysis and historical documentation. Since computer aided imaging techniques enable a quantitative assessment of the images under evaluation, it helps to improve the efficacy of the medical practitioners in arriving at an unbiased and objective decision in a short span of time. In addition, the use of multi-sensor [1] and multi-source image fusion methods offer a greater diversity of the features used for the medical analysis applications; this often leads to robust information processing that can reveal information that is otherwise invisible to human eye. The additional information obtained from the fused images can be well utilized for more precise localization of abnormalities.

The growing appeal of this research area can be observed from the large number of scientific papers published in the journals and magazines since year 2000 [2]. Figure 1 shows the increased frequency of publications in the field of medical image fusion from year 1995 to 2013. This can be largely attributed to the increased use of medical diagnostic devices by the medical community supported by rapid growth in low cost computing and imaging techniques. In addition to the rapid development of the imaging and computing technologies, there has been increased trust placed in diagnostics technologies in the medical field, as the new generation of medical practitioner finds the technologies user friendly. As opposed to the technologies that prevailed before the year 2000, the newer technologies pay a high level of attention on the usability and simplification of technical knowhow for the operator, making it friendly for the medical personnel. In addition, the services offered by the manufacturing companies have grown across in terms of geographic reach, speed and quality. These diverse set of factors has resulted in the medical imaging market annual growth rate of 7% and is expected to reach $49 billion by 2020. The applicability of imaging has shifted from just being a research tool to more towards a necessary diagnostic tool even in regular (non-research) hospitals across the world.

There exist several medical imaging modalities that can be used as primary inputs to the medical image fusion studies. The selection of the imaging modality for a targeted clinical study requires medical insights specific to organs under study. It is practically impossible to capture all the details from one imaging modality that would ensure clinical accuracy and robustness of the analysis and resulting diagnosis and. The obvious approach is to look at images from multiple modalities to make a more reliable and accurate assessment. This often requires expert readers and is often targeted at assessing details that complement the individual modalities. Some of the major modalities in clinical practice includes the following: (a) angiography such as Quantitative Coronary Angiography (QCA) and Quantitative Vascular Angiography (QVA), (b) Computer Tomography such as angiography (CTA), Quantitative Computed Tomography (QCT), (c) Dual-energy X-ray absorptiometry (DXA) such as for Bone Mineral Density (BMD) and Hip Structural Analysis (HSA), (d) Magnetic resonance imaging such as for Angiography (MRA), Body and Neuro, Cardiac, Dynamic Contrast-Enhanced Magnetic Resonance Imaging (DCE-MRI), (e) Nuclear Medicine such as using Multi-Gated Acquisition Scan (MUGA) and Single Photon Emission Computed Tomography (SPECT), (f) PET, (g) Ultrasound such as for Abdominal/small parts Ultrasound, Echocardiography, Intima-Media Thickness (IMT), Intravascular Ultrasound (IVUS), FMD (flow mediated dilation), Duplex, Duplex Doppler, CEUS (Contrast Enhanced Ultrasound), B-Mode and M-Mode, and (h) X-Ray imaging such as for mammography. These imaging modalities find a range of application in diagnosis and assessments of medical conditions effecting brain, breast, lungs, liver, bone marrow, stomach, mouth, teeth, intestines, soft tissues and bones.

The aim of this review is to provide a collective view of the applicability and progress of information fusion techniques in medical imaging useful for clinical studies [3, 4, 5, 6, 7, 8, 9, 10, 11, 12, 13, 14]. Figure 2 shows the three major focused areas of studies in medical image fusion: (a) identification, improvement and development of imaging modalities useful for medical image fusion, (b) development of different techniques for medical image fusion, and (c) application of medical image fusion for studying human organs of interest in assessments of medical conditions. There exist several image fusion studies that can be directly applied for fusing medical images. On a first look, this may seem a field that does not involve specific technical challenges and the reuse of image fusion algorithms for medical image rather a trivial task. However, the task of image fusion with medical images involve several technical challenges ranging from the limitations imposed by specific imaging modality, the nature of the clinical problem, the technology costs for the user, and the trust placed by the medical practitioners in the imaging technique. In the subsequent sections, we review these aspects of the medical image fusion, not just limiting to the grouping of medical image fusion algorithms, but also to categorizing the fusion techniques and applications based on the imaging modalities and organs of study.

## 2. Medical image fusion methods

Figure 3 shows the summary of the two stages involved in medical image fusion methods. The two stages of any classical image fusion method are (a) image registration and (b) fusion of relevant features from the registered images. The registration of the images requires a method to correct the spatial misalignment between the different image data sets that often involve compensation of variability resulting from scale changes, rotations, and translations. The problem of registration becomes complicated in the presence of inter-image noise, missing features and outliers in the images. On the other hand, the fusion of the features involve the identification and selection of the features with a focus on relevance of the features for a given clinical assessment purpose. Table 1 shows the summary of the major medical image fusion methods, the modalities that these methods are applied and the applications in medical imaging studies.

### 2.1. Morphological Methods

The morphology operators has been explored by image processing community for long, and the concept is used by the medical imaging community to detect spatially relevant information from the medical images. The morphological filtering methods for medical image fusion have been applied, for example, in brain diagnosis [47, 15, 49]. An example of modalities used in morphology based fusion can be seen in the fusion of CT and MR images [15, 16, 17]. In such applications, the morphology operators depend heavily on the structuring operator that defines the opening and closing operations. A calculated sequencing of the operations results in the detection of scale specific features. These features from different modalities can be used in for image fusion. The inaccuracies of detecting the features are high when the images are prone to noise and sensing errors. The operators such as averaging, morphology towers, K-L transforms and morphology pyramids are used for achieving the data fusion. These methods are highly sensitive to the inter-image variability resulting from outliers, noise, size and shape of the features.

### 2.2. Knowledge based methods

In medical imaging, there are several instances where the medical practitioner's knowledge can be used in designing segmentation, labeling and registration of the images. Generally, the domain-dependent knowledge is needed to set constraints on region-based segmentation and to make explicit the expectation of the appearance of the anatomy under the imaging modality at the stage of grouping the detected regions of interest. There are a range of applications where the domain-dependent knowledge is useful for image fusion such as for segmentation [18], micro-calcification diagnosis [19], tissue classification [20], brain diagnosis [20], classifier fusion [21], breast cancer tumor detection [21] and delineation & recognition of anatomical brain object [18]. The knowledge based systems can used in combination with other methods such as pixel intensity [19]. These methods place a significant amount of trust in the medical expert in labeling and identifying the domain knowledge relevant to the fusion task. The advantage is the ability to benchmark the images with the known human vision standards, while the drawback is the limitations imposed by human judgment in images that are prone to large pixel intensity variability. The use of preprocessing techniques in images can improve the imaging quality and increase the accuracy of ground truths.

### 2.3. Wavelet based methods

The primary concept used by the wavelet based image fusion [61, 26, 27, 32, 62, 63, 64, 65, 66, 40, 29, 30, 67, 68, 33, 69, 70, 71, 72, 32, 73, 74, 75, 59, 76, 77, 78, 79, 80, 34, 81, 82, 83, 84] is to extract the detail information from one image and inject it into another. The detail information in images is usually in the high frequency and wavelets would have the ability to select the frequencies in both space and time. The resulting fused image would have the "good" characteristics in terms of the features from both images that improve the quality of the imaging. There are several models for injection, the simplest being substitution. There exist several mathematical models for injection, such as simple addition operation and aggregator functions to more complex mathematical models. Irrespective of the models used, for practical reasons, the image resolution remains same before and after the fusion. In addition, the image resolution of the reference image enforces the required number of multiple levels of decomposition, such that a high resolution image would require more number of decomposition levels than a low resolution image. There are several applications of the wavelets in image fusion such as medical image pseudo coloring [85], super resolution [26], medical diagnosis [27, 28, 29, 30], feature level image fusion [31], lifting scheme [31], segmentation [32], 3D conformal radiotherapy treatment planning [33] and color visualization [34].

The feature level improvements on the images by combining wavelets with other techniques have proved to be useful for wavelet based image fusion. The most prominent approach of wavelet image fusion is with neural network [27, 28, 40], where the neural network often takes the roles of feature processing and wavelets take the role of a fusion operator. Similar to neural network, the kernel based operators such as support vector machines (SVM) can be used along with wavelets to achieve image fusion at feature levels [66]. Considering wavelets as a fusion operator, several feature processing methods can be combined such as wavelet-SVM [66], wavelet-texture measure [29], wavelet-MRA [30, 67], wavelet-self adaptive operator [69], wavelet-resolution-entropy [70, 72], nonlinear wavelet-shift invariant imaging [71], ICA-wavelet [86], wavelet-edge feature [75], wavelet-genetic [59], wavelet-contourlet transform [81], neuro-fuzzy-wavelet [82] and wavelet-entropy [84].

## 2.4. Neural Network based methods

Artificial neural networks (ANN) are inspired from the idea of biological neural network having the ability to learn from inputs for processing features and for making global decisions. The artificial neural network models require an input training set to identify the set of parameters of the network referred to as weights. The ability of the neural network models to predict, analyze and infer information from a given data without going through a rigorous mathematical solution is often seen as an advantage. This makes the neural network attractive to image fusion as the nature of variability between the images is subjected to change every time a new modality is used. The ability to train the neural network to adopt to these changes enable several applications for medical image fusion such as solving the problems of feature generation [36], classification [36], data fusion [36, 19, 27], image fusion [37, 38, 27, 39, 40, 41, 42, 43], micro-calcification diagnosis [19], breast cancer detection [38, 44, 45], medical diagnosis [27, 28, 42], cancer diagnosis [46], natural computing methods [87] and classifier fusion [45].

Although ANN offers generality in terms of having the ability to apply the concept of training, the robustness of ANN methods is limited by the quality of the training data and the accuracy of convergence of the training algorithm. In order to improve the quality of the features and thereby to improve the robustness of the ANN, hybrids of neural networks and sequential processing with other fusion techniques can be employed. Some of examples of these are wavelet-neural network [27, 28, 40], neural-fuzzy [41, 43], fuzzy-genetic-neural network-rough set [87] and SVM-ANN-GMM [45]. It is practically very difficult to prove the effectiveness of these combinations across all the different imaging modalities as these approaches are skewed towards the quality of the images selected for training that can vary significantly from one imaging condition to another.

## 2.5. Methods based on Fuzzy Logic

The conjunctive, disjunctive and compromise properties of the fuzzy logic have been widely explored in image processing and have proved to be useful in image fusion. The fuzzy logic is applied both as a feature transform operator or a decision operator for image fusion [47, 51, 48, 52, 53, 49, 54, 55, 50, 56, 41, 57, 58, 60, 88, 59, 87, 89, 90, 91, 43, 82, 92]. There are several applications of fuzzy logic base image fusion such as brain diagnosis [47, 48, 49, 50], cancer treatment [51], image segmentation and integration [51, 52], maximization mutual information [53], deep brain stimulation [54], brain tumor segmentation [55], image retrieval [56, 57], spatial weighted entropy [56], feature fusion [56], multimodal image fusion [41, 58, 59], ovarian cancer diagnosis [60], sensor fusion [88], natural computing methods [87] and gene expression [89, 90].

The selection of membership functions and fuzzy sets that result in the optimal image fusion is an open problem. The improvements of feature processing and analysis can be improved to fit the fuzzy space better when combined with probabilistic approaches such as fuzzy-neural network [41, 43], fuzzy-genetic-neural network-rough set [87], fuzzy-probability [89] and neuro-fuzzy-wavelet [82].

## 2.6. Other Methods

There are several methods that are based on dimensionality reduction techniques such as independent component analysis (ICA) [86, 93] and principal component analysis (PCA) [94, 95, 96, 97]. These dimensionality reduction techniques often find their use as feature processing methods and are used in combination with techniques such as ones based on wavelets [86]. A first order fusion of volumetric medical imagery is presented in [98]. A multimodal image fusion based on PCA using the intensity-hue-saturation (IHS) transform has been shown to preserve spatial features and required functional information without color distortion [97]. There are different mathematical transforms on features that can enhance the performance of the image fusion. For example, combination of complex contourlet transform with wavelet has been shown to result in robust image fusion [95, 96]. Transforms based methods are also applied for liver diagnosis [99], risk factor fusion [100], prediction of multifactorial diseases [100], parametric classification [100], local image analysis [101], multi-modality image fusion [102, 103, 95, 104, 96, 81]. Possibilistic clustering methods show improvement over the fuzzy c-means clustering and have a wide range of application in registration stages of image fusion. Some of the applications of possibilistic clustering include tissue classification [105], brain diagnosis [48, 106] and automatic segmentation [52]. SVM based techniques are kernel based techniques that are data and parameter driven having a strong control over the feature space. The ability of the SVM to reject the outliers in the data makes it a useful tool in image fusion, and leads to their being used in applications of cancer diagnosis [46, 107], classifier fusion [107, 108, 45], breast cancer tumor [108, 45], image fusion [66, 109], content-based image retrieval [110, 111], tumor segmentation [109], gene classification [112] and feature fusion [111]. Since SVMs can be used in registration as well as fusion stages, they can be combined with other methods to improve the speed of processing and accuracy when processing large image feature space under the influence of noise. Some examples of combined use of SVM with other methods include SVM-wavelet [66], SVM-adaptive similarity [110], SVM-data fusion [109] and SVM-ANN-GMM [45]. A prediction fusion is explained in [113]. Use of quaternionic signals representation for analysis and fusion of multi-components 2D medical images is presented in [114]. The use of ICA for the fusion of brain imaging data is presented in [115]. A Text fusion watermarking in medical image with semi-reversible for secure transfer and authentication is explained in [116]. Fusion of multiple expert annotations for medical image diagnosis is reported [117]. Fast fusion of medical images based on Bayesian risk minimization and pixon map is presented [118].

## 3. Imaging modalities used in image fusion

Figure 3 shows a few examples of image fusion with different medical imaging modalities. In this example, the image fusion is achieved in MRI-PET fusion [119] using a contourlet method, MRI-SPECT [119] uses a wavelet based approach, MRI-CT [82] uses a Integer Wavelet Transform and Neuro-Fuzzy method, PET-CT [120] uses a wavelet coefficients fusion method, and Vibroacoustography images with X-ray mammography [121] uses a linear combination approach.

### 3.1. Magnetic Resonance Imaging

Magnetic Resonance Imaging (MRI) plays an important role in non-invasive diagnosis of brain tumors and is one of the most widely used imaging modalities in medical studies in trusted clinical settings. Previous work [122, 16, 61, 123, 124, 26, 125, 53, 126, 127, 64, 128, 129, 130, 131, 91, 132, 133, 134] reports the successful fusion of MR images with different types of modalities. The image fusion methods are widely applied for brain diagnosis and treatment [47, 15, 135, 48, 20, 136, 137, 49, 138, 139, 50, 140, 141, 142, 143, 144], wherein the fusion techniques have been demonstrated to show improved imaging and diagnostic performances.

Image segmentation is widely used to identify objects and regions of interest in the images. In MR medical imaging, the most common use of segmentation is for extraction of the different types of tissues and to identify abnormal regions such as reflective of tumors. Several tumor segmentation methods are reported [18, 52, 123, 126, 32, 50, 145, 109, 51, 144, 146, 147] that aid to improve the accuracy of tumor identification and automatic detection of tumors from the MR images. The segmentation along with medical image fusion techniques have been widely used in prostate localization [148, 149, 150, 151, 152, 153, 154, 155, 156, 157, 158, 159, 160, 161, 162, 163, 164, 165, 166, 146, 131, 167, 168, 169, 170, 171, 133].

There are several clinical visualization applications such as 3D conformal radiation therapy [148, 172] that employ fusion techniques. Following along similar lines, medical image visualization method [127], and its extension to adaptive medical image visualization system, is developed using the hierarchical neural network and intelligent decision fusion [36]. The fusion techniques are even extended to 3D voxel fusion in a view to achieve multi-modality image fusion [173].

The MRI based image fusion also finds application in prostate studies. Image fusion techniques [174, 150, 151, 175, 155, 156, 158, 165, 176] are extensively used in prostate seed implant quality assessment. The fusion techniques are even applied to develop stereotactic prostate biopsy system [177]. There are methodological advancements in the fusion of MR images such as structure similarity match measure (SSIM) [178] that can improve the accuracy of these applications. Other potential applications that incorporate MRI based fusion include image regeneration [179, 180, 143], potential field visualization [181], lung/liver diagnosis [182], tissue classification [105, 20], breast cancer assessment [183, 184, 185], surgical planning and training [186], multi-dimensional visualization [124], extraction of shape, color and structure of specimen [187], visualization and pattern recognition [188], image registration [125, 152, 53, 127, 189, 132], MRI guided treatment [152, 153], gynecological cancer diagnosis [190] and 3D tumor simulation [191].

The advantage of MRI is that it is very safe for pregnant women and babies as it does not involve any exposure to radiation. In addition, the soft issue structures in organs such as brain, heart and eyes are imaged with high accuracy. The major disadvantage of the MRI images is its relative sensitivity to movement, making it a difficult technique for assessing organs that involve movement such as with mouth tumors. The use of image fusion can overcome this limitation in a multi-modal imaging environment, enabling reconstruction and prediction of the missing information from MRI. The MR images along with other modalities when used together with modern image fusion techniques have shown to improve the imaging accuracy, and practical clinical applicability. There exists several studies that attempt to combine the MRI with other modalities using image fusion methods, some examples of this are the following: MRI-CT-PET-SPECT-DSA-MEG [47, 135], MRI-CT [15, 16, 17, 148, 174, 179, 180, 192, 193, 61, 150, 187, 194, 151, 173, 53, 175, 155, 156, 172, 157, 158, 159, 189, 64, 195, 196, 161, 197, 178, 164, 198, 142, 165, 129, 91, 167, 168, 199, 133, 176], MRI-CT-PET [51, 191, 200, 201, 202], EEG-MRI [181], CT-FOCAL [182], MRI-Mammogram [183, 185], MR-SPECT [48, 124, 188, 152, 203, 137, 141], MRI-SPECT-PET [48], nuclear medicine- MRI [204], endoscopy-MRI [186], MRI-CT-SPECT [26, 205], MRI-Molecular [136], MRA-DSA [125], MRI/CT-PET-SPECT [190], MRI-iMRI-SPECT [153], MRI-PET [206, 207, 208, 140, 184, 209, 130, 131], MRI-DTI [210, 145, 143], ultrasound-MRI [211, 149, 162, 170, 177], and MRI-TRUS [212, 166, 146, 169, 171], MRI-MRSI [163]. The most prominent combination is the MRI-CT studies largely because of the maturity in the technology and practical usability in clinical settings.

### 3.2. Computerized Tomography

Computerized tomography (CT) is a medical imaging technique that has made a prominent impact on medical diagnosis and assessments. This is a popular modality used in multi-modal medical image fusion [16, 61, 213, 26, 214, 215, 64, 216, 29, 217, 68, 218, 129, 76, 91, 133]. Similar to MR images, the CT images are used in a vast range of medical applications under practical clinical conditions. Computerized assessment using CT images has been one of the early attempts towards modern medical imaging; an example system is one that uses knowledge-based image interpretation system for the segmentation and labeling of a series of 2-D brain X-ray CT-scans [18]. Another application of the image fusion using CT images has been for assistance in surgical planning, training and guidance using images obtained from a tracked endoscope to surfaces derived from CT data [186].

The CT images have gain importance as a 3D imaging technique, and image fusion has been applied in applications such as those that use 3D tumor simulations [191]. Similar to MRI, the application of CT images in brain diagnosis and treatment [47, 15, 135, 142] has also been reported. There are several application areas where CT images are considered the prime modality. some of these major applications are the following: (1) head and neck cancer diagnosis [219, 192, 220, 221, 222, 199], (2) cancer treatment [51, 223], (3) image segmentation and integration [51, 216], (4) lung cancer treatment [224, 225, 182, 226, 227, 228, 229, 230, 231, 232], (5) prostate cancer treatment [148, 233, 234, 150, 235, 151, 154, 155, 156, 236, 237, 157, 158, 238, 159, 239, 160, 216, 128, 240, 164, 241, 165, 167, 168, 242, 133], (6) 3D conformal radiation therapy [148, 227, 172, 232], (7) liver diagnosis [99, 182], (8) prostate seed implant quality assessment [174, 150, 151, 175, 155, 156, 158, 243, 176], (9) image regeneration [179, 180], (10) tumor detection [244, 245, 246, 220, 247, 247], (11) extraction of shape, color and structure of specimen [187], (12) gynecological cancer diagnosis [190], (13) 3D Voxel fusion [173], (14) gross tumor volume detection [200, 248], (15) pelvic irradiation treatment [236, 238, 249], (16) diagnosis of local recurrence of rectal cancer [214], (17) colorectal cancer treatment and chemotherapy [250], (18) pediatric solid extracranial tumors [251], (19) deep brain stimulation [195], (20) bone tumor surgery [196], (21) telemedicine [201], (22) localization [240], (23) breast cancer assessment [217, 252, 253], (25) vulvar cancer treatment [205], (26) oral cancer treatment [254], (27) bone cancer diagnosis [255], (28) lung cancer diagnosis [256], (29) radiation therapy and planning [230, 164], (30) biopsy [257], (31) cervical cancer treatment [202], (32) orbital tumor surgery [129], (33) liver tumor diagnosis [258], classification fusion [259], (34) esophageal cancer diagnosis [260], and (35) pancreatic tumors characterization [261].

The main advantages of the CT scan are the relative short scan times and high imaging resolutions. However, the exact radiation levels are not well understood topics, and CT has several other limitations such as limited tissue characterization because of the nature of X-ray probe, restriction of CT scan to transverse slices and practical limitation on number of X-rays that can be produced in the short scan times. Fusion combinations in which CT is one of the main modalities include MRI-CT-PET-SPECT-DSA-MEG [47, 135, 164], MRI-CT [15, 16, 17, 148, 174, 179, 180, 262, 61, 150, 187, 151, 263, 173, 154, 175, 155, 156, 172, 157, 158, 159, 189, 64, 195, 160, 196, 128, 161, 178, 198, 142, 165, 129, 91, 167, 168, 199, 242, 133, 176], SPECT-CT [219, 224, 99, 244, 236, 250, 238, 216, 217, 257, 218, 247, 255], MRI-CT-PET [51, 191, 200, 201, 202], CT-FOCAL [182], ultrasound-CT [264, 234, 237, 241], FDG-CT [226, 227], nuclear medicine-CT [204], endoscopy-MRI [186], MRI-CT-SPECT [26, 205], MRI/CT-PET-SPECT [190], CT/SPET-SRS [245], FDG-PET-CT [265, 229, 251, 253], PET-CT [248, 266, 214, 246, 220, 239, 29, 223, 221, 68, 254, 256, 252, 222, 76, 259, 267, 261], TRUS-CT [268], Ultrasound-CT [269, 243, 258], PET-CT-ultrasound [260].

### 3.3. Positron Emission Tomography

Positron emission tomography, widely known as PET imaging or a PET scan, is a useful type of nuclear medicine imaging. Here, we discuss some application areas where PET is a prime modality considered in the data fusion. Similar to CT and MRI, a major application of PET is in radiology studies for brain diagnosis and treatment [47, 135, 48, 208, 140, 143]. There are a wide range of the application of image fusion using PET, some of which are for cancer treatments [51, 223, 220, 162, 222, 209, 225, 229, 131], image segmentation and integration [51, 259], 3D tumor simulation [191], gynecological cancer diagnosis [190], inertial electrostatic confinement fusion [270], gross tumor volume detection [200, 271], diagnosis of local recurrence of rectal cancer [214], tumor detection and treatment [246, 220], pediatric solid extracranial tumors [251], telemedicine [201], breast cancer detection [184, 252, 253], oral cancer treatment [254], lung cancer diagnosis [256, 272], cervical cancer treatment [202], esophageal cancer diagnosis [260], and pancreatic tumors characterization [207, 261].

The resolution limits of PET image are one of the main challenges. There is often an integrated approach to reduce the limitations by modeling finite resolution effects in image reconstruction, and improved detector design. The high sensitivity provided by the molecular imaging is often seen as an advantage of PET images. There is increased interest in using fusion techniques to improve the imaging quality. The use of PET data in combination with some of the existing modalities using the image fusion techniques include MRI-CT-PET-SPECT-DSA-MEG [47, 135], MRI-CT-PET [51, 191, 200, 201, 202], MRI-SPECT-PET [48], MRI/CT-PET-SPECT [190], MRI-PET [207, 208, 140, 184, 209, 143, 130, 131], FDG-PET-CT [229, 251, 253], PET-CT [273, 248, 214, 246, 220, 221, 239, 29, 223, 68, 254, 256, 252, 222, 76, 274, 261], FDG-PET [272], and PET-CT-ultrasound [260].

### 3.4. Single-Photon Emission Computed Tomography

Single photon emission computed tomography (SPECT) scan is useful nuclear imaging method that is widely used to study the blood flow to tissues and organs. The application areas include brain diagnosis and treatment [47, 135, 48, 137, 141], head and neck cancer diagnosis [219, 203], lung cancer treatment [224], liver diagnosis [99], tumor detection [244], fusion of multi-modality images [124, 26, 53, 239, 216, 218], multi-dimensional visualization [124], visualization and pattern recognition [188], fMRI guided treatment [152, 153], prostate cancer treatment [153, 236, 238, 239, 216], gynecological cancer diagnosis [190], image registration [53], pelvis irradiation treatment [236, 238], colorectal cancer treatment and chemotherapy [250], breast cancer assessment [217], vulvar cancer treatment [205], bone cancer diagnosis [255], and biopsy [257].

Improving the sensitivity of the SPECT without reducing the image resolution is one of the main challenges in SPECT imaging. The developments in pin-hole SPECT is used to enhance the resolution capabilities to sub millimeter range. The imaging quality is however still affected by the signal noise, and improving the image quality and resolution requires post-processing techniques. The image fusion with PET attempts to improve the imaging quality and includes PET-CT [275], PET-MRI-CT [276, 277], MRI-CT-PET-SPECT-DSA-MEG [47, 135], SPECT-CT [219, 224, 99, 244, 236, 250, 238, 239, 216, 217, 257, 255], MR-SPECT [48, 124, 188, 125, 53, 203, 137, 141, 218], MRI-CT-SPECT [26, 205], MRI/CT-PET-SPECT [190] and MRI-iMRI-SPECT [153].

### 3.5. Ultrasound

Ultrasound imaging is sonar based imaging technique that has been used widely due to its low cost and no known side effects to the patients. There is a wide range of applications where the ultrasound images are used to infer medical data. Some of these are for prostate cancer treatment [271, 149, 278, 234, 279, 237, 162, 164, 241, 170], conformal radiation therapy [271], brachytherapy prostate implant [264, 278, 234, 237, 164], image fusion [279, 61, 162], breast cancer detection [280], liver tumor diagnosis [258], prostate biopsy [177], and esophageal cancer diagnosis [260].

There are some major limitations of the ultrasound imaging that are tightly linked to the operator skills, such as the need to ensure no air gaps between the probe and body, and the need to avoid bone structures in the path of organ imaged. These major deficiencies necessitate the need to use other modalities to ensure the accuracy of imaging and localization of the regions under test for diagnostic measurements. Examples of fusion techniques that incorporate ultrasound in it are ultrasound-X-rays [271], ultrasound-CT [264, 234, 237, 164, 241, 258], ultrasound-uoroscopic [278], nuclear medicine-ultrasound [204], microscopy-ultrasound [37], ultrasound-CAD-mammograms-infrared [280], ultrasound-MRI [149, 162, 170, 177] and PET-CT-ultrasound [260].

### 3.6. Other Modalities

Along with the modalities mentioned in the previous subsections, there are several other imaging methods such as infrared, fluorescent, microwave and microscopic imaging that find application through medical image fusion. Infrared as an imaging modality is used in the application of breast cancer detection [38, 280]. A fusion combination that includes infrared can be seen in [280] where the combination consists of ultrasound, CAD, mammograms and infrared images. Fluorescent imaging has been used as an application to oral cancer detection [281] and prostate brachytherapy and treatment [278, 282, 283]. The image fusion of fluorescent images with other modalities can be found between ultrasound-fluoroscopic [278] and TRUS-fluoroscopic images [282]. Microwave imaging is used in breast cancer detection [284, 285] and tumor identification [284]. Another modality to point out is the microscopic imaging used in image fusion [37, 286]. Microscopic imaging is used in fusion methods as an application to image mosaicing [286], multi-feature fusion, feature extraction, and global/local recognition [287]. A feedback retina model for improving medical images fusion is presented in [288]. In [289], an attempt to use information fusion in medical decision support systems is presented. There have been successful attempts to apply image fusion techniques with microscopy and ultrasound images [37]. Trans-rectal ultrasound (TRUS) is a variant of ultrasound imaging that is used in prostate brachytherapy dosimetry [282, 151, 166, 146, 169, 171], image guided prostate intervention [290], biopsy planning [212, 169, 171], segmentation [146], and prostate seed implant quality assessment [243]. Fusion combinations with TRUS with other imaging modalities can be found in TRUS-uroscopic [282], TRUS-MRI [290, 212, 166, 146, 169, 171] and TRUS-CT [243]. Mammography is an X-ray based imaging modality that has been widely used for breast cancer assessment [183, 280, 185], micro-calcification diagnosis [19] and image registration [291]. The image fusion of mammogram with other modalities can significantly improve the detection accuracies of problems such as abnormal tissue identification in case of calcification. There exist several combinations of modalities with mammograms such as MRI-mammogram [183], ultrasound-CAD-mammograms-infrared [280] and MR Mammogram -X mammogram [136]. A tool for medical image fusion and visualization is presented in [292]. Another modality which is gaining popularity is molecular imaging and image fusion in application to brain diagnosis and treatment [136] has shown to improve the imaging interpretations. An example of the image fusion with molecular imaging is in combination with MRI [136].

## 4. Major application domains (organs

### 4.1. Brain

Brain is one of the important organs that have been subjected to a wide range of medical image analysis and research. The imaging studies reveal several important pieces of information about the brain which are otherwise not visible to human sensory mechanisms. The most commonly used image modalities to study the brain include CT [47, 15, 16, 17, 18, 135, 186, 293, 294, 142, 259], MRI [47, 15, 16, 17, 85, 181, 105, 135, 48, 52, 186, 124, 20, 136, 137, 126, 49, 208, 295, 54, 139, 55, 210, 93, 294, 141, 142, 144, 296], DSA [47], PET [47, 135, 48, 208, 259], SPECT [47, 135, 48, 124, 137, 141], MEG [47], EEG [181], Endoscopy [186], Molecular [136] and DTI [210].

Medical image fusion in brain studies has been employed for segmentation of brain tissues [18, 105, 52, 126, 295, 139, 55, 50, 210, 144, 259, 296], pseudo coloring for MRI based brain segmentation [85], visualizing cortical potential fields [181], stereotactic brachytherapy of brain tumors [135], brain tissue map and volume identification [105], image guided neuro-surgery [186, 297], development of stereoscopic panoramas of brain images[186], 2D-3D registration of brain images [186, 136, 208], volumetric fusion with brain images [124], classification of abnormal brain tissues [20], semiautomatic 3D fusion with brain images[136], image fusion with multimodal brain images [298, 136, 293, 208, 195, 210, 141], verification of implanted catheters [293], surface projection maximum mutual information fusion of brain images [137], parametric classification of differential brain activity [299], locating anatomical targets with MRI brain images[295], microelectrode recording and test stimulation [195], multi-classifier fusion based brain image segmentation [139], classification of differential brain activity [300], sensor fusion for surgical navigation [301], emulation of perceptual system of brain[ 302], ensemble based data fusion for diagnosis of Alzheimer's disease [303], filter bank selection for brain computer interaction [304], feature based fusion of brain images [93], optic chiasm contouring for monitoring of brain tumors[294], brain tumor biopsy [305], multimodal fusion of muscles and brain signals [306], and decoding visual brain states [307, 308].

### 4.2. Breast

The breast has been subject of several studies due to the high rates of breast cancer in women. The most commonly used modality for breast studies is mammogram (both analogue and digital), followed by MRI and/or CT. The combinations of PET (functional imaging) and X-ray computed tomography (CT, anatomical localization) has shown significant improvements in diagnostic accuracy, allowing better differentiation between normal (e.g. bowel) and pathological uptake. The modalities that has been used to study breast include MRI [183, 184, 185], Mammogram [183, 280, 185], Infrared [38, 280], Ultrasound [280], Microwave [284, 285], PET [184, 252], SPECT [217, 257] and CT [252, 257]. The image fusion applications targeted on breast include prediction breast cancer tumors [21, 108, 44], breast surgery [309], ultrawideband breast cancer detection [285], and breast cancer detection in premenopausal women [310].

### 4.3. Prostate

Prostate is another organ that has been studied using multi-modal medical images. There exists a range of techniques and studies on prostate based image fusion, that often face the challenge deformation of prostate in multi-modal imaging setups [271, 148, 149, 174, 264, 278, 234, 279, 282, 150, 151, 152, 311, 155, 312, 156, 236, 237, 215, 159, 239, 160, 313, 162, 163, 241, 165, 168, 169, 171]. The medical image analysis techniques on prostate include localization of the prostate for 3D conformal radiation therapy [148], prostate localization in the post-planning setting [314] evaluation of prostate gland motion and volume changes [315], evaluation of prostate seed brachytherapy [234], thermal ablation of the prostate cancer [152,153], histology study of prostate tissue [279], post implant dosimetric analysis of prostate brachytherapy [157, 161], prostate brachytherapy [237, 157, 161], and biomechanical modeling of prostate motion [316]. The imaging modalities that deal with prostate include ultrasound [271, 149, 264, 278, 234, 279, 237, 162, 241, 316], X-rays [271], CT [148, 174, 264, 234, 150, 154, 239, 155, 156, 236, 237, 157, 160, 313, 161, 241, 165, 168], MRI [148, 149, 174, 150, 153, 154, 155, 156, 157, 160, 161, 162, 163, 165, 168, 169, 171], uroscopic [278, 282], fMRI [152, 153], SPECT [152, 153, 236, 313], PET [239, 160] and TRUS [282, 169, 171].

### 4.4. Lungs

Lung is a vital organ that undergoes direct contact with environment through the air intake and is the main part of respiratory system. Lungs are prone to damage from pollutants and viruses. The imaging of the lungs can often reveal several details that reflect the condition of the internal tissues. The ability to distinguish a damaged tissue, cancerous tissue and a healthy tissue is not an easy task in early diagnosis. Image fusion techniques have been shown to improve the diagnostic performance and screening, and especially improve the clinical monitoring outcomes [224, 225, 182, 226, 227, 229, 256, 230, 231, 272, 232, 317]. Example of medical image analysis on lungs includes localization study on potentially operable non-small cell lung cancer [225] and dosimetric planning for non–small-cell lung cancer [317]. There exists several modalities that is applied in lungs studies such as SPECT [224], PET-CT [275], FDG-PET [318], CT [224, 225, 182, 226, 227, 229, 256, 230, 231, 232], PET [225, 226, 229, 256, 272] and FDG [226, 227, 229, 272].

### 4.5. Other Organs

Liver is another vital organ that is being increasingly studied using images, and the complexity of the liver tissue makes the medical imaging studies challenging. The registration and fusion of liver images for medical diagnosis is a task of primary importance [99]. The major modalities that deal with liver studies include SPECT [224], CT [224, 182, 319, 258], PET [319] and ultrasound [258]. In bone marrow imaging, the medical diagnosis that performs image fusion includes tumor cell identification in bone marrow [320], extraction of bone shape, color and structure of bone specimen [187] and bone tumor surgery [196, 321]. MRI [187, 196] and CT [187, 196, 321] are the modalities that are used for bone marrow imaging. On pelvis, image fusion methods are used for gynecological cancer diagnosis [190, 322] and analysis of conformal pelvic irradiation [236]. Ovarian cancer diagnosis uses fuzzy rule base classifier fusion [60]. The use of secondary data to estimate instantaneous model parameters of diabetic heart disease is explained in [323]. A semantic based fusion technique in application to Alzheimer's disease is presented in [324]. A fusion imaging using a hybrid SPECT-CT camera is used in colorectal [250] cancer patients. The imaging of gross tumor volume delineation in head and neck cancer also investigated in the past [246, 220]. Detection of oral cancer using fluorescent image by color image fusion is also explored [281].

## 5. Discussions and Conclusions

The field of medical diagnostics and monitoring using medical images faces several technological, scientific and societal challenges. The technological advancements in imaging technologies have resulted in improved imaging accuracies. However, every modality of imaging has its own practical limitations, which is further imposed by the underlying nature of the organ and tissue structures. This enforces the need to explore the possibility to newer imaging technologies and to explore the possibility of using multiple imaging modalities. The ability of image fusion techniques to quantitatively and qualitatively improve the quality of imaging features makes multi-modal approaches efficient and accurate relative to unimodal approaches. The availability of a large number of techniques in feature processing, feature extraction and decision fusion makes the field of image fusion appealing to be used by medical imaging community. The methodological innovations specific to medical image fusion algorithms is rather limited at this stage, as majority of medical image fusion algorithms are derived from existing image fusion studies. The main challenge in applying image fusion algorithms is to ensure the medical relevance and aid for a better clinical outcome. The right combination of the imaging modalities, feature processing, feature extraction and decision fusion algorithms that targets a specific clinical problem in itself is a challenging and nontrivial task. Even the same images under consideration often require very different types of processing for different types of diagnostics over a region of interest. The major issues concerning feature processing and extraction algorithms resulting from the presence of pixel intensity outliers, missing features, sensors errors, spatial inaccuracies, and inter-image variability remain an open problem in medical image fusion. The inaccurate registration of the objects between the images is tightly linked to the poor performance of feature or decision level fusion on medical image fusion algorithms, and requires medical domain knowledge and algorithmic insights to reduce the fusion inaccuracies.

Another, point of interest is that when addressing the medical image fusion problems, the emphasis has been in the direction of developing algorithms that try to improve the imaging quality and regions of interest within images. The need for improving the image quality arises from the signal noise and the physical limitations of the imaging modality. The estimation of signal noise and compensation is considered as an important problem in medical imaging, and the advancements in enhancements to image quality can have a positive impact on the image fusion process. Another area of interest is to improve the speed of processing especially in the cases of volumetric image fusion. An algorithmic approach is to develop algorithms that are optimized for high speed processing. However, they would be limited by the hardware and operating system capabilities. An alternate approach is to develop real-time processing systems in field programmable gate arrays and dedicate parallel computing graphical processing units. The speed is of primary importance in real-time image fusion during surgery or that involve continuous real-time monitoring. These are emerging areas of thoughts, and would require substantial progress in image fusion systems research.

The progress of this field largely depends on the trust that the medical practitioner and medical institutions place on the clinical improvements resulting from medical image fusion approaches. This is not an easy task, and would require a substantial convincing effort through technology improvements, access to the technological advancements and improving the usability of multi-modal systems in clinical setup. There are several technological advancements that can propel this growth. The primary growth comes from low-power high performance computing hardware developments for imaging that can process large volumes of high resolution images. In many medical imaging applications, although image resolutions are very high, the existing limitation in computing hardware makes the processing of such images in a time-limited clinical setup impossible. The advancements in parallel computing hardware such as low cost graphical processing units can overcome much of the problems facing conventional algorithmic approaches. The development of low cost computing also depends on the advancements in the semiconductor technologies and how quickly the technologies can be transferred to the market. The development in cognitive computing algorithms and hardware is another major technological advancement that could have a significant impact in the way in which the images are processed and presented. The incorporation of natural learning techniques in imaging hardware and software would be the natural progression that would aim to compete with human judgment - which by far would be the most challenging aspect to adopt in the medical service industry, but an obvious technological advancement for progressing medical image fusion research.

In conclusion, image fusion techniques in terms of medical image modalities and organs of study have been discussed in this survey. The extensive developments in medical image fusion research summarized in this literature review indicate the importance of this research in improving the medical services such as diagnosis, monitoring and analysis. The availability and growth of a wide range of imaging modality has enabled progress in medical image fusion to be useful for clinical deployment. Although, there has been significant progress in the medical image fusion research, the application of the general fusion algorithms is limited by the practical clinical implications as imposed by the medical experts based on the requirements of specific medical studies. In addition to medical reasons, there exists technical challenges in image registration and fusion resulting from image noise, resolution difference between images, inter-image variability between the images, lack of sufficient number of images per modality, high cost of imaging and increased computational complexity with increasing image space and time resolution. Nonetheless, even under these challenging situations, the fused images provide the human observers improved viewing and interpretation of medical images. The algorithms used for medical image fusion studies have resulted in the improved imaging quality and have proved to be useful for clinical applications. The prominent approaches include wavelets transforms, neural networks, fuzzy logic, morphology methods, and classifiers such as support vector machines. Combining one or more image fusion methods is also observed to be successful in medical image analysis. The algorithmic approaches to image fusion are also limited by the imaging hardware. The development of equipment that can perform multi-modal scanning is a challenging topic as it involves the risk of exposing the patients to additional radiation, longer examination time, and increased cost of the device. This also involves having to look at compatibility issue of technologies as the space-time resolution and scanning speeds vary substantially from one imaging modality to another. The problem is much more significant in developing image fusion algorithms and devices for real-time medical applications such as robotic guided surgery. Since several of these challenges remain open and the image fusion in medical imaging has proved to be useful and the trust in these techniques is on the rise, it is expected that the innovation and practical advancements would continue to grow in the upcoming years.

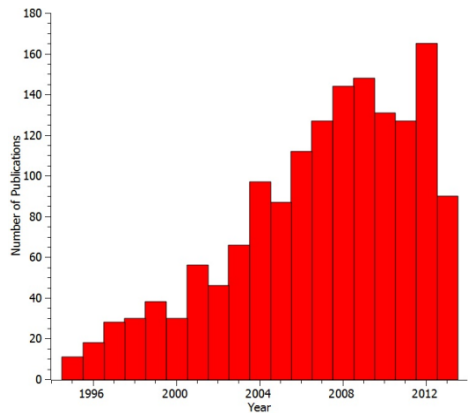

Figure 1: The frequency of publications in medical image fusion as obtained from the ISI knowledge of web indexing service.

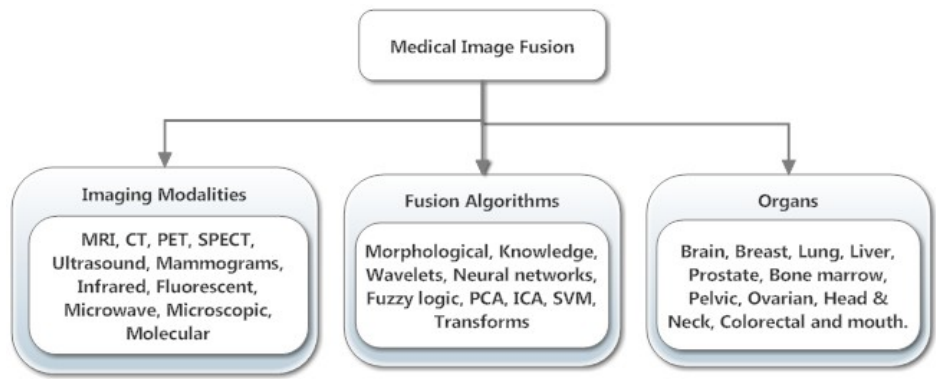

Figure 2: A chart showing the nature of modalities, methods and organs of interest as applied in medical image fusion studies .

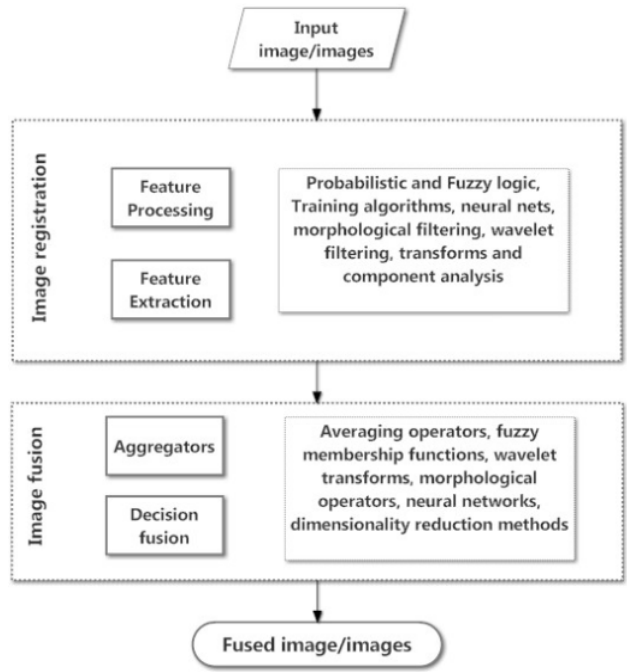

Figure 3: The summary of the stages in the image fusion of medical images. The two stage process consists of image registration followed by image fusion.

| Combination | Modality 1 | Modality 2 | Fused Image |
|---|---|---|---|
| MRI-PET | 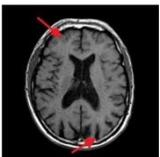 | 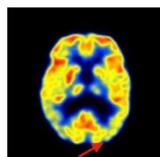 | 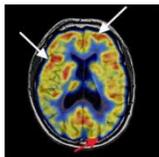 |
| MRI-SPECT | 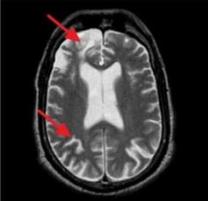 | 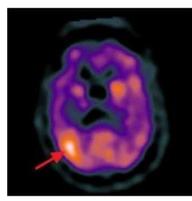 | 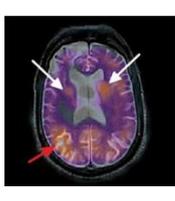 |
| MRI-CT | 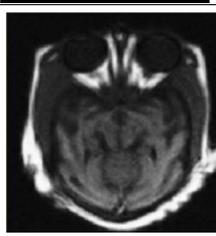 | 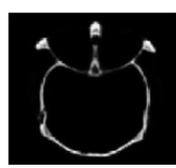 | 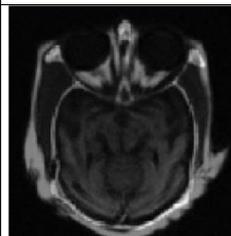 |
| Xray-VA | 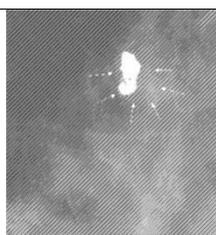 | 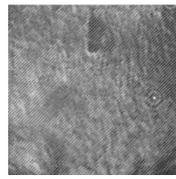 | 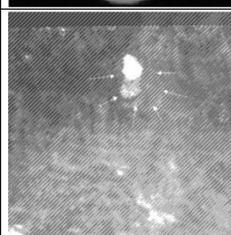 |
| PET-CT | 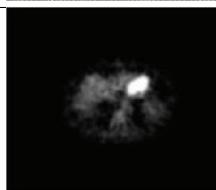 | 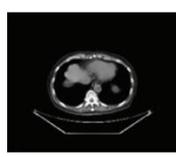 | 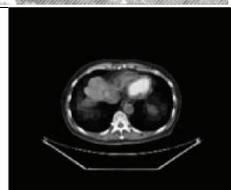 |

Figure 4: Examples of multi-modal medical image fusion. the combination of modality 1 with modality 2 using specific image fusion techniques results in improved feature visibility for medical diagnostics and assessments as shown in the fused image column.

Table 1: Major medical image fusion methods and its applications

| Method | Modalities | Applications | Fusion strategies |
|---|---|---|---|
| Morphology | MRI, CT | Brain diagnosis [15, 16, 17] | Morphology filters and pyramids |
| Knowledge | MRI, CT, PET, Ultrasound, Mammogram | Segmentation [18], micro-calcification diagnosis [19], tissue classification [20], brain diagnosis [20], classifier fusion [21], breast cancer tumor detection [21, 22], delineation & recognition of anatomical brain object [18] and medical image retrieval [23, 24, 25] | Knowledge learning systems, expert systems |
| Wavelets | CT, PET, MRI, Mammograms, MRA, fMRI, SPECT, ultrasound | pseudo coloring, super resolution [26], medical diagnosis [27, 28, 29, 30], feature level image fusion [31], lifting scheme [31], segmentation [32], 3D conformal radiotherapy treatment planning [33] and color visualization [34] | Wavelet transform, multi resolution analysis [35], Discrete wavelet transform, Stationary wavelet transform, dual tree discrete wavelet transform, Lifting wavelet transform, Multi-wavelet transform Coupled |
| Artificial neural networks | CT, PET, MRI, Mammograms, MRA, fMRI, SPECT, ultrasound | feature generation [36], classification [36], fusion [36, 19, 27, 37, 38, 27, 39, 40, 41, 42, 43], micro-calcification diagnosis [19], breast cancer detection [38, 44, 45], medical diagnosis [27, 28, 42], | neural networks, clustering neural network, fuzzy neural networks, wavelet neural networks |

| | | cancer diagnosis [46] | |
|---|---|---|---|
| Fuzzy logic | CT, PET, MRI, Mammograms, MRA, fMRI, SPECT, ultrasound | Brain diagnosis [47, 48, 49, 50], cancer treatment [51], image segmentation and integration [51, 52], maximization mutual information [53], deep brain stimulation [54], brain tumor segmentation [55], image retrieval [56, 57], spatial weighted entropy [56], feature fusion [56], multimodal image fusion [41, 58, 59], ovarian cancer diagnosis [60] | Image fuzzification, modification of membership values, Image defuzzification, fuzzy combination operators, neuro-fuzzy networks |